\pdfoutput=1
\documentclass[11pt]{article}

\usepackage[]{acl}

\usepackage{times}
\usepackage{latexsym}

\usepackage[T1]{fontenc}
\usepackage[utf8]{inputenc}

\usepackage{microtype}

\usepackage{makecell}
\usepackage{todonotes}
\usepackage{colortbl,xcolor}
\usepackage{tikz, pgfplots}
\usepackage{graphicx}
\usepackage{booktabs}
\usepackage{amssymb}
\usepackage{amsmath}
\usepackage{multirow}
\usepackage{float}
\usepackage{setspace}
\usepackage{cancel}
\usepackage[normalem]{ulem}

\usepackage{pifont}
\newcommand{\cmark}{\ding{51}}%
\newcommand{\xmark}{\ding{55}}%

\title{System-Initiated Transitions from Chit-Chat to Task-Oriented Dialogues with Transition Info Extractor and Transition Sentence Generator}

\author{Ye Liu$^1$$^,$$^3$, Stefan Ultes$^2$, Wolfgang Minker$^3$ \and Wolfgang Maier$^1$ \\
$^{1}$Mercedes-Benz AG, Sindelfingen, Germany \\
\texttt{\{ye.y.liu, wolfgang.mw.maier\}@mercedes-benz.com}\\
$^{2}$University of Bamberg, Bamberg, Germany \\
\texttt{stefan.ultes@uni-bamberg.de}\\
$^{3}$Ulm University, Ulm, Germany \\
\texttt{\{ye.liu, wolfgang.minker\}@uni-ulm.de}}

\begin{document}
\maketitle

%\iffalse

\begin{abstract}

%In chit-chat human-machine interaction, a dialogue system should be capable of detecting if the user requires some task-oriented service (like booking a train in Figure \ref{fig: system-initiated transition from chit-chat to task-oriented.}) at a proper moment. This shows proactivity and is also beneficial for commercial dialogue systems to actively sell their services. 

In this work, we study dialogue scenarios that start from chit-chat but eventually switch to task-related services, and investigate how a unified dialogue model, which can engage in both chit-chat and task-oriented dialogues, takes the initiative during the dialogue mode transition from chit-chat to task-oriented in a coherent and cooperative manner.
%build an initiative conversational model that can proactively guide this transition to task-related services through generating a transition sentence.
We firstly build a \emph{transition info extractor} (TIE) that keeps track of the preceding chit-chat interaction and detects the potential user intention to switch to a task-oriented service. Meanwhile, in the unified model, a \emph{transition sentence generator} (TSG) is extended through efficient Adapter tuning and transition prompt learning. When the TIE successfully finds task-related information from the preceding chit-chat, such as a transition domain (``train'' in Figure \ref{fig: system-initiated transition from chit-chat to task-oriented.}), then the TSG is activated automatically in the unified model to initiate this transition by generating a transition sentence under the guidance of transition information extracted by TIE. The experimental results show promising performance regarding the proactive transitions. We achieve an additional large improvement on TIE model by utilizing Conditional Random Fields (CRF). The TSG can flexibly generate transition sentences while maintaining the unified capabilities of normal chit-chat and task-oriented response generation. 
%The traditional conditional random fields (CRF) technology highly improves the performance of TIE. %The proposed two different types of prompts enable flexibility of TSG.

\end{abstract}

\section{Introduction}
\label{sec: introduction}

\begin{figure*}[!ht]
\centering
\footnotesize
\scalebox{0.95}{
\begin{tikzpicture}

\node [] at (0, 0) {\includegraphics[width=.03\textwidth]{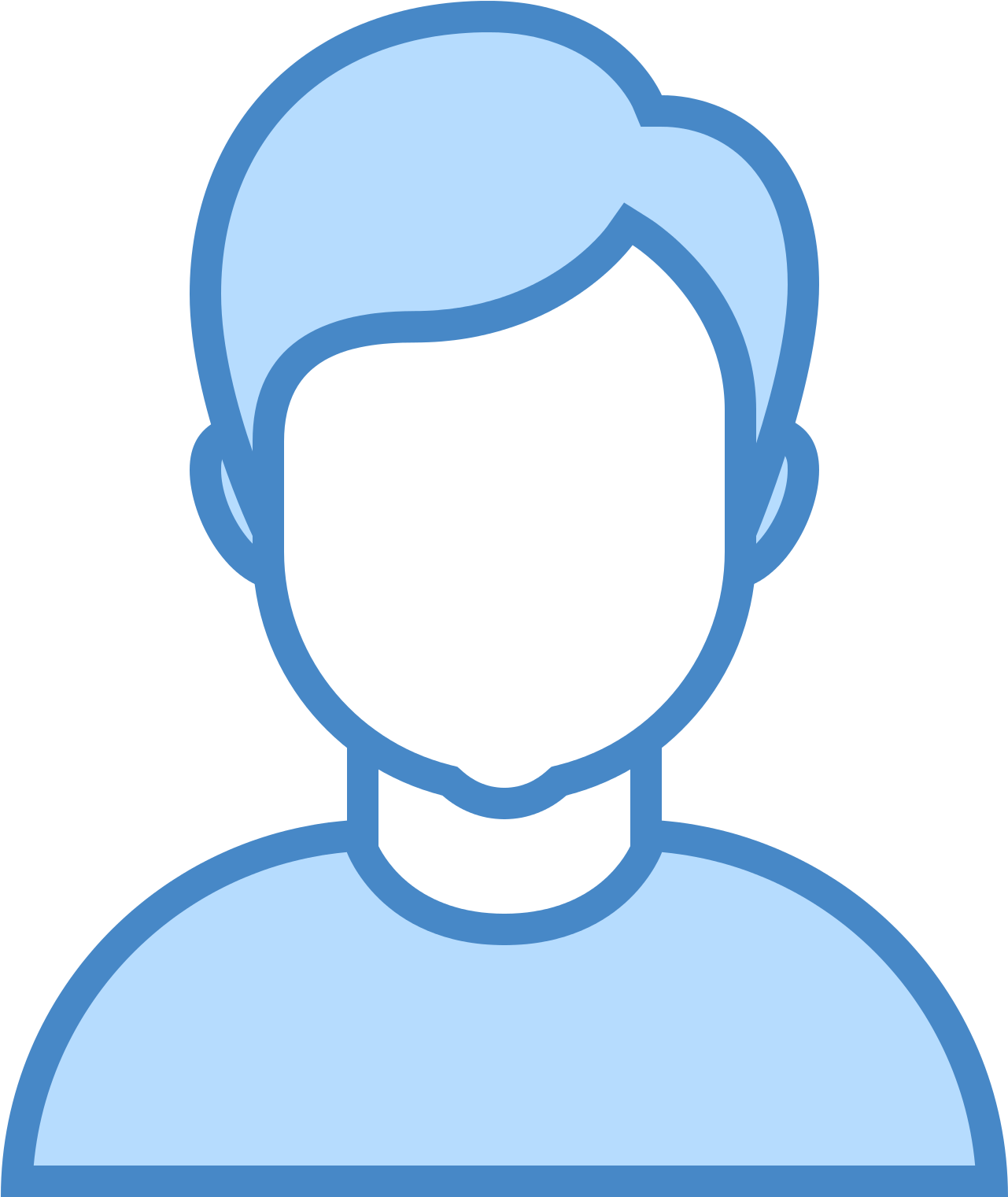}};
\draw [rounded corners] (0.5, -0.2) rectangle (11.5, 0.2);
\node [] at (6.0, 0.0) {\textcolor{cyan}{I will be enrolling in a new school at \textbf{London Kings Cross} next week. I'm so nervous.}};

\node [] at (15, -0.6) {\includegraphics[width=.04\textwidth]{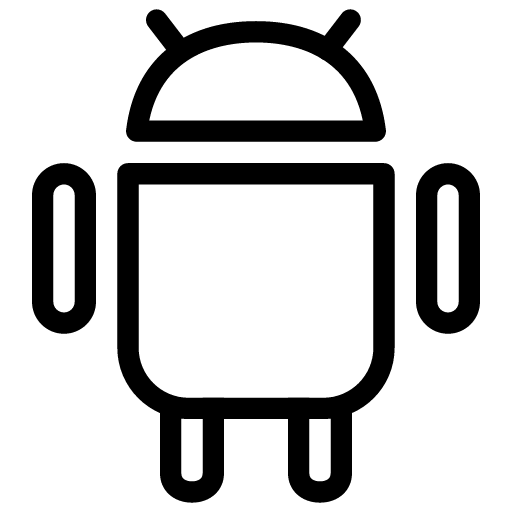}};
\draw [rounded corners] (9.0, -0.4) rectangle (14.5, -0.8);
\node [] at (11.8, -0.6) {\textcolor{cyan}{I hope you have fun at your new school.}};

\node [] at (0, -1.2) {\includegraphics[width=.03\textwidth]{pictures/user.png}};
\draw [rounded corners] (0.5, -1.0) rectangle (12, -1.4);
\node [] at (6.2, -1.2) {\textcolor{cyan}{Thank you. My family and I will be visiting my school this weekend to see how it's like.}};

\node [] at (15, -2.1) {\includegraphics[width=.04\textwidth]{pictures/robot.png}};
\draw [rounded corners] (12.5, -1.6) rectangle (14.5, -2.0);
\node [] at (13.5, -1.8) {\sout{\textcolor{cyan}{I see.}}};
\draw [rounded corners] (4.5, -2.1) rectangle (14.5, -2.5);
\node [] at (9.5, -2.3) {\textcolor{cyan}{I see.} \textcolor{red}{If you want, I can look for a train to London Kings Cross for you.}};
%\textcolor{magenta}{Additionally I could help with looking for train tickets for you.} }};
%\node [] at (10.0, -2.2) {// \textcolor{red}{If you want, I can look for a train to London Kings Cross for you.}};

\node [] at (0, -2.9) {\includegraphics[width=.03\textwidth]{pictures/user.png}};
\draw [rounded corners] (0.5, -2.7) rectangle (13.0, -3.1);
\node [] at (6.7, -2.9) {\textcolor{orange}{I'm trying to find a train that goes from Cambridge to my school. Can you help me book a ticket?}};

\node [] at (15, -3.4) {\includegraphics[width=.04\textwidth]{pictures/robot.png}};
\draw [rounded corners] (5.5, -3.2) rectangle (14.5, -3.6);
\node [] at (10.0, -3.4) {\textcolor{orange}{I can help with that. Can you tell me what day you will be travelling?}};

\draw[dotted,thick,black] (7.5, -3.7) -- (7.5, -3.8);

\end{tikzpicture}
}
\caption{A Prepended FusedChat dialogue with an augmented transition sentence (\textcolor{red}{red}) for the proactive transition from chit-chat to task-oriented.
The \textcolor{cyan}{blue} and \textcolor{orange}{orange} represents chit-chat and task-oriented interaction respectively. Compared with the original chit-chat (crossed out) response at the transition turn, the transition sentence (\textcolor{red}{red}) can enable the dialogue system to proactively switch to task-oriented services.
}
\label{fig: system-initiated transition from chit-chat to task-oriented.}
\end{figure*}

Spoken dialogue systems (SDSs) have usually been developed targeting only one out of two different categories, task-oriented or chit-chat (aka open-domain). The former focuses on achieving functional goals and the latter aims at creating engaging social conversations without special goals. 
In recent years, several previous works \citep{lin2021adapter, zhao2021unids, young2022fusing} have studied unified conversational models that can engage in both chit-chat and task-oriented dialogue.
%In recent years, creating a unified conversational model \citep{lin2021adapter, zhao2021unids, young2022fusing} that can engage in both chit-chat and task-oriented dialogue has aroused great interest.
However, the system-initiated transitions that emerge during switchover between these two dialogue modes have rarely been explored.
Especially when a user chats casually with the dialogue system, but implicitly expresses a need for a specific task-related service,
it is desired that the dialogue system is able to capture this hidden information and proactively ask the user if they require such a task-oriented service (like booking a train ticket in Figure \ref{fig: system-initiated transition from chit-chat to task-oriented.}).
It has been proven to be beneficial for commercial SDSs to proactively offer or sell their task-related services \citep{chiu2022salesbot, liu2023unified}.
Furthermore, these transitions smoothly initiated by the dialogue system are regarded as a proactive feature \citep{2014fn02} and can greatly improve the user interaction experience \citep{liu-etal-2022-system}.

%Two transition sentences are attached here and split with {//}. The former \textcolor{magenta}{magenta} one is only domain (train) guided transition sentence, the latter \textcolor{red}{red} one is the transition domain-slot-value (train-destination-\textcolor{red}{London Kings Cross}) guided transition sentence.

The goal of this work is to develop the initiative capabilities of a unified conversational model that is capable of detecting the implicit user intention of using some task-related services, even if they are talking casually, and to proactively bridge the connection from chit-chat to task-oriented dialogue through generating a transition sentence (\textcolor{red}{red} in Figure \ref{fig: system-initiated transition from chit-chat to task-oriented.}). As the dialogue example in Figure \ref{fig: system-initiated transition from chit-chat to task-oriented.} shows, the original response at the transition turn is only ``I see''. If the agent can anticipate in advance that the user wants to visit the ``London Kings Cross'' through the preceding chit-chat, it can then proactively establish a connection with the task-oriented ``train'' service that the user needs by saying ``If you want, I can look for a train to London Kings Cross for you.''.

To enable the initiative capabilities in a unified model, the main contributions of this paper are as follows:
\begin{enumerate}
    \item
    %We incorporate natural language understanding (NLU), which is generally an indispensable component in task-oriented SDSs for intent classification and slot filling \citep{chen2019bert}, into this work and build a transition info extractor (TIE) to keep track of the preceding chit-chat interaction meanwhile detecting the potential user intention switching to task-oriented services. 
    
    To detect the hidden task-related transition domain/slot/value entities from the preceding chit-chat, we propose the transition info extractor (TIE) to keep track of preceding chit-chat dialogue through leveraging natural language understanding (NLU) technology \citep{chen2019bert}.
    
    \item
    We artificially augment $215$ dialogues with a domain guided transition sentence and a domain-slot-value guided transition sentence respectively. We then collect transition sentence templates for different domains and different domain-slot pairs from these human augmented dialogues. The transition sentence templates are further utilized to annotate the remaining unannotated dialogues.

    \item
    We leverage transition prompt learning \citep{li2022personalized} and Adapter tuning \citep{lin2021adapter} to efficiently extend the transition sentence generation (TSG) in a unified NLG model with the augmented dialogues. 
\end{enumerate}

%we build a transition info extractor (TIE) to track the preceding chit-chat and detect the potential task-related transition domain and slot-value pair. Meanwhile, a unified Natural Language Generation (NLG) model is extended to proactively guide this transition by generating a transition sentence (\textcolor{red}{red} or \textcolor{magenta}{magental} in Figure \ref{fig: system-initiated transition from chit-chat to task-oriented.}) at a proper timing.
%The NLU model includes two classifiers, which are responsible for predicting transition domain and transition slot; and a Name Entity Recognition (NER) model to extract the transition value from the chit-chat dialogue history.

\begin{figure*}[!ht]
\centering
\footnotesize
\scalebox{1.0}{
\begin{tikzpicture}

\node [] at (0, 0.0) {{preceding chit-chat}};

\draw[->, thick] (1.5, 0.0) -- (2.0, 0.0);

\draw[thick, rounded corners] (2.2, -0.6) rectangle (3.2, 0.6);
\node at (2.7, 0.0)[align=center]  {{TIE}};

\path[->, red] (3.3, 0.1) edge [bend left] node { \makecell[c]{transition domain \cmark \\ transition slot \cmark \\ transition value \cmark } } (6.1, 0.1);

\path[->, dashed] (3.3, -0.1) edge [bend right] node { {transition info \xmark} } (6.1, -0.1);

\draw[thick, rounded corners] (6.2, -0.6) rectangle (7.2, 0.6);
\node at (6.7, 0.0)[align=center]  {\makecell[c]{unified \\ NLG}};

\draw[thick, red] (7.3, 0.8) rectangle (8.3, 1.2);
\node at (7.8, 1.0)[align=center]  {\makecell[c]{\textcolor{red}{TSG}}};

\path[->, red] (7.2, 1.0) edge [bend right] node {} (6.7, 0.7);

\draw[->, red, thick] (7.5, 0.4) -- (8.0, 0.4);
\node [] at (10.5, 0.4) {\textcolor{red}{normal response w/ transition sentence}};

\draw[->, thick] (7.5, -0.4) -- (8.0, -0.4);
\node [] at (10.5, -0.4) {\makecell[c]{normal response w/o transition sentence \\ (chit-chat or task-oriented)}};

\end{tikzpicture}}
\caption{\label{fig: architecture flow for the system-initiated transition from chit-chat to task-oriented} The overall Architecture flow for system-initiated transitions from chit-chat to task-oriented.}
%The \textcolor{red}{red} flow means that current is a good timing to make the initiative transition and the NLU model successfully extracts task-oriented transition information, like transition domain-slot-value. Then these information is further fed into the unified NLG model to trigger the generation of the transition sentence besides the normal response and proactively guide the dialogue switching to task-oriented interaction. When no transition information is available, the NLG only generate normal response, chit-chat or task-oriented response without transition sentence.
\end{figure*}
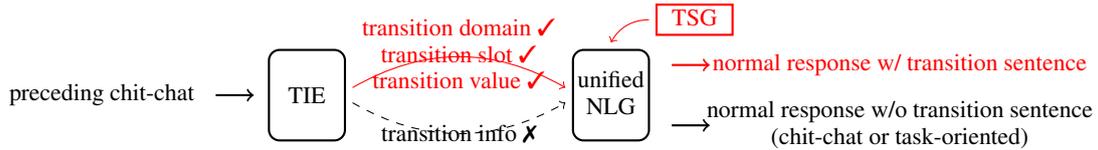

The overall architecture flow of this work is shown in Figure \ref{fig: architecture flow for the system-initiated transition from chit-chat to task-oriented}. When the TIE successfully extracts the transition information from the preceding chit-chat, the TSG in the unified NLG is activated to generate a transition sentence besides the normal response to proactively guide this switch. The combined flow is highlighted in \textcolor{red}{red}. Otherwise, the TIE continually tracks the chit-chat, and unified NLG works as usual to generate a normal chit-chat or task-oriented response without (w/o) a transition sentence.
%i.e., a response without transition sentence.

The remainder of this paper is structured as follows: Section~\ref{sec: related works} shows related work of our research. Section \ref{sec: Transition Sentence Augmentation and Templates} presents the transition sentence augmentation and templates for the TSG training. Section \ref{sec: NLU+CRF} introduces the proposed TIE model for detecting the task-related transition information from the preceding chit-chat interaction. Section \ref{sec: unified GPT2 and transition sentence adapter} presents the unified NLG extended with TSG through transition prompt and Adapter tuning. Section \ref{sec: results comparison} elaborates on the performance evaluation of this work. Section \ref{sec: conclusion} concludes this work and outlines future research.

\section{Related Works}
\label{sec: related works}

%Task-oriented SDSs aim at achieving functional goals in a modular framework. They are usually comprised of different components: Automatic Speech Recognition (ASR), NLU, Dialog Management (DM), NLG and Text-To-Speech (TTS). Among them, 

NLU is generally a crucial component in task-oriented SDSs and responsible for parsing an utterance into a semantic frame to identify the user’s intention \citep{de2008spoken}. With the development of deep learning methods, RNN, CNN, as well as their variations or combinations have been widely for the NLU task \citep{yao2013recurrent, mesnil2014using, yao2014spoken, liu2016attention}. \citet{wang2018attention} proposed a attention-based encoder-decoder, CNN-BLSTM, for joint intent detection and slot filling. \citet{goo2018slot} proposed a slot gate that focused on capturing the relationship between slot and intent. \citet{kenton2019bert} and \citet{xu2020end} both used the pre-trained BERT for the joint intent classification and slot filling. The proposed TIE is inspired by NLU modeling.

Beyond that, \citet{xu2013convolutional} and \citet{ma2016end} both utilized the traditional approach, Conditional Random Fields (CRF) \citep{sha2003shallow}, for sequence labelling with the combination of LSTM and CNN. We also leverage the CRF technology to further improve the performance of the TIE model.

\citet{shuster2020dialogue} introduced the dodecaDialogue task, to assemble important aspects of an engaging conversational agent into a single collection by leveraging $12$ tasks. Adapter-Bot \citep{lin2021adapter} utilized multiple adapter layers with the pre-trained DialoGPT model to activate new response skills and styles. \citet{zhao2021unids} proposed a dialogue model for training chit-chat and task-oriented in a unified data schema, which both include belief states, representation of dataset results, and system acts. However, these models simply fuse chit-chat dialogue and task-oriented dialogue into one model and do not consider the dependency between different types of dialogues in the multi-turn setting. 
In contrast, all dialogues in FusedChat released in \citet{young2022fusing} include both chit-chat and task-oriented turns, and treat them as parallel dialogue modes of equal importance. \citet{chiu2022salesbot} proposed SalesBot and introduced the dialogue transitions from chit-chat to task-oriented. \citet{liu2020towards} introduced the proactive transitions in conversational recommendation over multi-type dialogues.
\citet{liu-etal-2022-system} elaborated on three types of system-initiated transitions in a unified dialogue model and discussed the potential challenges respectively. 
\citet{liu2023unified} proposed the system-initiated transitions between chit-chat and task-oriented dialogues, where the transitions from chit-chat to task-oriented and from task-oriented to chit-chat were treated equally.
However, we mainly investigate the system-initiated transitions from chit-chat to task-oriented with the Prepended FusedChat dataset for this work.

\section{Transition Sentence Augmentation and Templates}
\label{sec: Transition Sentence Augmentation and Templates}

This section introduces the details of human augmentation of transition sentences and template collection for unannotated dialogues.

We mainly utilize the \textbf{Prepended} FusedChat \citep{young2022fusing} dataset for initiative transitions from chit-chat to task-oriented in this work.
FusedChat is a public available dataset, where human augmented open-domain dialogues are prepended and appended to the dialogues of the popular task-oriented dataset MultiWOZ \cite{budzianowski2018multiwoz, DBLP:journals/corr/abs-2104-00773}.
In the Prepended FusedChat, each dialogue starts with chit-chat interaction and eventually switch to task-oriented requests. Table \ref{tab: statistic of Prepended FusedChat} shows the statistics of the Prepended FusedChat\footnote{The FusedChat used in this work is the first version uploaded by author Young and and has minor differences to the current version of FusedChat available at https://github.com/tomyoung903/FusedChat.} used in this work. As a prepended FusedChat example shown in Figure \ref{fig: system-initiated transition from chit-chat to task-oriented.}, the user controls the switch to task-oriented services.\footnote{This is common in most of prepended FusedChat dialogues, as confirmed by manual analysis.}
However, our goal is to build a proactive dialogue system that can establish a smooth transition from chit-chat to task-oriented by itself.

To achieve this, we hire one master student with computational linguistics background to augment a domain guided transition sentence and a domain-slot-value guided transition sentence (\textcolor{red}{red} sentence in Figure \ref{fig: system-initiated transition from chit-chat to task-oriented.}) for $215$ Prepended FusedChat dialogues respectively. The domain guided transition sentence must explicitly include the domain information. The domain-slot-value guided transition sentence must contain the specific value extracted from the preceding chit-chat dialogue aside from the domain, as the transition sentence in Figure \ref{fig: system-initiated transition from chit-chat to task-oriented.}, ``If you want, I can look for a train to London Kings Cross for you.'' with ``train'' domain and ``London Kings Cross'' value.

%In the FusedChat dataset, human annotated open-domain dialogues were prepended or appended to the dialogues of the popular task-oriented dataset MultiWOZ \cite{budzianowski2018multiwoz, DBLP:journals/corr/abs-2104-00773}. Every dialogue in FusedChat includes two dialogue modes with the chit-chat and task-oriented parts being interdependent. In our work, we mainly use the Prepended FusedChat, where the prepended open-domain dialogue must retain a slot value mentioned in the first turn of the succeeding task-oriented dialogue to enable the inter-mode dependency. 

\begin{table}
\centering
\footnotesize
\scalebox{1.0}{
\begin{tabular}{cccc}
    \toprule
        data type & train  & valid & test \\
        dialogue size & 3255 & 474 & 331  \\
    \bottomrule
\end{tabular}}
\caption{Statistics of Prepended FusedChat.}
\label{tab: statistic of Prepended FusedChat}
\end{table}

%%%%%%%%%%%%%%%%%%%%%%%%%%
\begin{table}
\centering
\footnotesize
\scalebox{0.92}{
\begin{tabular}{cccccc}
    \toprule
         domain & train & restaurant & attraction & taxi   \\
         number of templates & 95 & 56 & 45 & 17 \\
    \bottomrule
\end{tabular}
}
\caption{Statistics of transition sentence templates for different domains.}
\label{tab:The statistic of transition sentence templates for different domains.}
\end{table}

%%%%%%%%%%%%%%%%%%%%%%%%%%
\begin{table*}
\centering
\footnotesize
\scalebox{0.95}{
\begin{tabular}{cccccccccc}
    \toprule
    
         domain & \multicolumn{3}{c}{train} & \multicolumn{2}{c}{restaurant} & \multicolumn{2}{c}{attraction} & \multicolumn{2}{c}{taxi} \\ 
         \cmidrule(lr{.75em}){2-4}  \cmidrule(lr{.75em}){5-6} \cmidrule(lr{.75em}){7-8}  \cmidrule(lr{.75em}){9-10}

         slot & day & destination & departure & food & name & type & name & destination & departure \\
    \midrule
         number of templates & 22 & 40 & 35 & 45 & 11 & 30 & 15 & 9 & 8 \\
    \bottomrule
\end{tabular}}
\caption{Statistics of transition sentence templates for different domain-slot pairs.}
\label{tab:The statistic of transition sentence templates for different domain-slot pairs.}
\end{table*}

After the human augmentation, we collect the templates for transition sentences in different domains and different domain-slot pairs from the augmented $215$ dialogues respectively. For the domain-slot-value guided transition sentences, we use ``[VALUE]'' to replace the specific value to collect the domain-slot templates. Table \ref{tab:The statistic of transition sentence templates for different domains.} and Table \ref{tab:The statistic of transition sentence templates for different domain-slot pairs.} show template statistics for different domains and domain-slot pairs, respectively. Table \ref{tab: transition sentences templates for different domains} and Table \ref{tab: transition sentences templates for different domain-slot pairs} in the Appendix show some template examples of transition sentences in different domains and domain-slot pairs respectively. These templates are further used to randomly annotate the remaining unannotated Prepended FusedChat dialogues. Then all Prepended FusedChat with augmented transition sentences can be used for training the extended TSG in the unified NLG.

%%%%%%%%%%%%%%%%%%%%%%%%%%
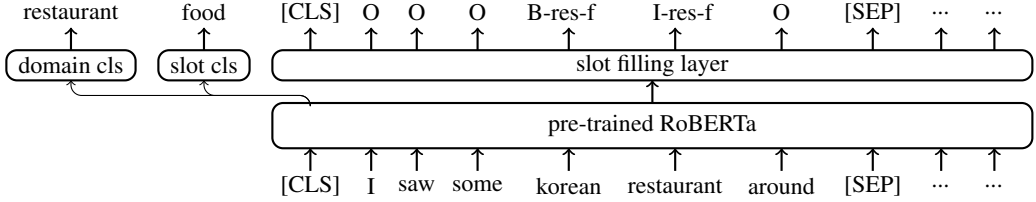
\begin{figure*}
\centering
\footnotesize
\scalebox{1.0}{
\begin{tikzpicture}

\draw[thick,rounded corners]   (-5.5, -0.3) rectangle (4.5, 0.3);
\node at (-0.5, 0)[align=center]  {{ pre-trained RoBERTa }};

\draw[->,thick] (-5.0, -0.6) -- (-5.0, -0.3);
\node at (-5.0, -0.8)[align=center]  {{ [CLS] }};

\draw[->,thick] (-4.2, -0.6) -- (-4.2, -0.3);
\node at (-4.2, -0.8)[align=center]  {{ I }};

\draw[->,thick] (-3.6, -0.6) -- (-3.6, -0.3);
\node at (-3.6, -0.8)[align=center]  {{ saw }};

\draw[->,thick] (-2.8, -0.6) -- (-2.8, -0.3);
\node at (-2.8, -0.8)[align=center]  {{ some }};

\draw[->,thick] (-1.6, -0.6) -- (-1.6, -0.3);
\node at (-1.6, -0.8)[align=center]  {{ korean }};

\draw[->,thick] (-0.2, -0.6) -- (-0.2, -0.3);
\node at (-0.2, -0.8)[align=center]  {{ restaurant }};

\draw[->,thick] (1.2, -0.6) -- (1.2, -0.3);
\node at (1.2, -0.8)[align=center]  {{ around }};

\draw[->,thick] (2.4, -0.6) -- (2.4, -0.3);
\node at (2.4, -0.8)[align=center]  {{ [SEP] }};

\draw[->,thick] (3.3, -0.6) -- (3.3, -0.3);
\node at (3.3, -0.8)[align=center]  {{ ... }};

\draw[->,thick] (4.0, -0.6) -- (4.0, -0.3);
\node at (4.0, -0.8)[align=center]  {{ ... }};

\draw[->,thick] (-0.5, 0.3) -- (-0.5, 0.6);
\draw[thick,rounded corners]   (-5.5, 0.6) rectangle (4.5, 1.0);
\node at (-0.5, 0.8)[align=center]  {{ slot filling layer }};

\draw[->,thick] (-5.0, 1.0) -- (-5.0, 1.3);
\node at (-5.0, 1.5)[align=center]  {{ [CLS] }};

\draw[->,thick] (-4.2, 1.0) -- (-4.2, 1.3);
\node at (-4.2, 1.5)[align=center]  {{ O }};

\draw[->,thick] (-3.6, 1.0) -- (-3.6, 1.3);
\node at (-3.6, 1.5)[align=center]  {{ O }};

\draw[->,thick] (-2.8, 1.0) -- (-2.8, 1.3);
\node at (-2.8, 1.5)[align=center]  {{ O }};

\draw[->,thick] (-1.6, 1.0) -- (-1.6, 1.3);
\node at (-1.7, 1.5)[align=center]  {{ B-res-f }};

\draw[->,thick] (-0.2, 1.0) -- (-0.2, 1.3);
\node at (-0.1, 1.5)[align=center]  {{ I-res-f }};

\draw[->,thick] (1.2, 1.0) -- (1.2, 1.3);
\node at (1.2, 1.5)[align=center]  {{ O }};

\draw[->,thick] (2.4, 1.0) -- (2.4, 1.3);
\node at (2.4, 1.5)[align=center]  {{ [SEP] }};

\draw[->,thick] (3.3, 1.0) -- (3.3, 1.3);
\node at (3.3, 1.5)[align=center]  {{ ... }};

\draw[->,thick] (4.0, 1.0) -- (4.0, 1.3);
\node at (4.0, 1.5)[align=center]  {{ ... }};

\draw[->,rounded corners] (-5.0, 0.3) -- (-5.0, 0.4) -- (-6.4, 0.4)-- (-6.4, 0.6);
\draw[thick,rounded corners]   (-7.0, 0.6) rectangle (-5.8, 1.0);
\node at (-6.4, 0.8)[align=center]  {{ slot cls }};
\draw[->,thick] (-6.4, 1.0) -- (-6.4, 1.3);
\node at (-6.4, 1.5)[align=center]  {{ food }};

\draw[->,rounded corners] (-5.0, 0.3) -- (-5.0, 0.4) -- (-8.15, 0.4)-- (-8.15, 0.6);
\draw[thick,rounded corners]   (-9.0, 0.6) rectangle (-7.3, 1.0);
\node at (-8.15, 0.8)[align=center]  {{ domain cls }};
\draw[->,thick] (-8.15, 1.0) -- (-8.15, 1.3);
\node at (-8.15, 1.5)[align=center]  {{ restaurant }};

\end{tikzpicture}
}
\caption{\label{fig: transition domain and slot prediction, slot filling in the proposed NLU task} Architecture of the proposed TIE model that includes transition domain/slot classifier and slot filling task. The ``B-res-f'' and ``I-res-f'' is the abbreviation of the ``B-restaurant-food'' and ``I-restaurant-food'' respectively.
%Only one chit-chat utterance is attached because of the space limitation. However, the TIE model will keep track of all the preceding chit-chat.
}
% Besides predicting transition domain and corresponding slot, the slot filling task is proposed to extract the specific value from dialogue history, like the ``korean restaurant''. When it is not a good moment for transition, then the predicted domain and slot should be ``UNK'' and slot filling label is all with ``O''.
\end{figure*}

\section{Transition Info Extractor (TIE)}
\label{sec: NLU+CRF}

This section presents our TIE model that can detect potential user intention to switch to task-oriented services. As shown in Figure \ref{fig: transition domain and slot prediction, slot filling in the proposed NLU task}, TIE is adapted from pre-trained RoBERTa \citep{liu2019roberta} and has three components, a transition domain classifier, a transition slot classifier and a slot filling layer.
When the interaction starts from chit-chat, the TIE keeps track of the preceding chit-chat to predict the potential transition domain and slot, while extracting the specific value through slot filling. For instance, the transition domain-slot-value extracted in Figure \ref{fig: transition domain and slot prediction, slot filling in the proposed NLU task}, is ``restaurant-food-Korean restaurant''.

\subsection{Joint RoBERTa for domain/slot classification and slot filling}
\label{subsec: Joint RoBERTa for domain/slot classification and slot filling}

We utilize the pre-trained RoBERTa \citep{liu2019roberta} as the backbone TIE model for jointly predicting transition domain and corresponding slot, also extracting the specific value from the preceding chit-chat dialogue through slot filling task, as shown in Figure \ref{fig: transition domain and slot prediction, slot filling in the proposed NLU task}.

Given the preceding dialogue history until to the current user turn $\mathbf{x} = (\emph{x}_{1}, \emph{x}_{2}, ... \emph{x}_{n})$, the [CLS] token is inserted into the first place and [SEP] is inserted to split user utterances and system responses. The corresponding slot filling label is $\mathbf{y}^{\textit{sf}} = (\emph{y}_{1}, \emph{y}_{2}, ... \emph{y}_{n})$ along with [CLS] and [SEP] tokens. The input $\mathbf{x}$ and slot filling label $\mathbf{y}$ are both padded to maximal length $N$ of the batch data. In addition, $\emph{y}^{d}$ and $\emph{y}^{s}$ are transition domain and slot label respectively. Let $\mathcal{D} = \{ (\mathbf{x}, \emph{y}^{d}, \emph{y}^{s}, \mathbf{y}^{\textit{sf}}) \}_{m=1}^{M}$ be the dataset of size $M$ for joint RoBERTa training.

Adapted from the pre-trained RoBERTa, the final hidden states of the input are 
\begin{equation}
\label{equ: RoBERTa output}
    \emph{h}_{\text{[CLS]}}, \emph{h}_{x_{1}}, \emph{h}_{x_{2}}, ..., \emph{h}_{x_{n}} = \text{RoBERTa}(\mathbf{x})
\end{equation}
Two classifier layers in Equation \ref{equ: RoBERTa domain/slot classifiers} are separately added on the output of [CLS] token, $\emph{h}_{\text{[CLS]}}$, to predict transition domain and slot.
\begin{equation}
\label{equ: RoBERTa domain/slot classifiers}
\begin{aligned}
    \hat{\emph{y}}^{d} & = \textit{softmax} (\mathbf{W}^{d} \textit{Dropout}(\emph{h}_{\text{[CLS]}}) + \mathbf{b}^{d}) \\
    \hat{\emph{y}}^{s} & = \textit{softmax} (\mathbf{W}^{s} \textit{Dropout}(\emph{h}_{\text{[CLS]}}) + \mathbf{b}^{s}) \\ 
\end{aligned}
\end{equation}

For the domain classifier, four different transition domains,\footnote{``hotel'' also exists in Prepended FusedChat as transition domain, but only in two dialogues. We delete those two dialogues to prevent the severe imbalance between different domains.} train, restaurant, attraction and taxi, are collected in the Prepended FusedChat. When no explicit user intention is detected in the preceding chit-chat, the domain classifier should recognise it as ``UNK'' to indicate that the current dialogue turn is not a good moment to switch to task-oriented.
Hence, the domain classifier is a $5$ classification task. 

For the slot classifier, there are six slots,\footnote{Two dialogues have ``pricerange'' as transition slot under restaurant domain and one dialogue has ``area' as transition slot under attraction domain. We also remove these dialogues in case of the imbalanced slots.} namely day, destination, departure, food, name and type. Also along with ``UNK'', the slot classifier is a $7$ classification task. Some slots are shared in different domains, e.g., ``name'' in restaurant and attraction domains (see Table \ref{tab:The statistic of transition sentence templates for different domain-slot pairs.}).

For the slot filling task, the final hidden states in Equation \ref{equ: RoBERTa output} are fed into the slot filling ($\textit{sf}$) layer in Equation \ref{equ: RoBERTa slsot filling} to classify over slot filling labels. 
\begin{equation}
\label{equ: RoBERTa slsot filling}
    {\hat{\mathbf{y}}_{n}^{\textit{sf}}} = \textit{softmax} (\mathbf{W}^{\textit{sf}} \textit{Dropout}(\mathbf{h_{x_{n}}}) + \mathbf{b}^{\textit{sf}} ) ; \\
        n \in 0 ... N \\
\end{equation}
We use the IOB (In/Out/Begin) labelling format \citep{ramshaw1999text} for the slot filling labels. The dictionary of those labels is as follows and includes $22$ tokens:
\begin{itemize}
    \item
    $3$ special tokens, ``[PAD]'', ``[CLS]'', ``[SEP]'', which are aligned with RoBERTa tokenizer.
    %mapping to the ``[PAD]'', ``[CLS]'' and ``[SEP]'' of RoBERTa tokenizer.
    
    \item
    $9$ domain-slot combinations in Table \ref{tab:The statistic of transition sentence templates for different domain-slot pairs.}, but every domain-slot pair is extend with prefix ``B-'' and ``I-''. E.g. ``B-restaurant-food'' and ``I-restaurant-food'' in the Figure \ref{fig: transition domain and slot prediction, slot filling in the proposed NLU task}. When the specific value has more than one word, the first one is labelled with prefix ``B-'', the remaining with prefix ``I-''.
    
    \item
    The ``O'' is assigned to words not belonging to any specific value in sentences.
\end{itemize}

The $\mathbf{W}$ and $\mathbf{b}$ in Equation \ref{equ: RoBERTa domain/slot classifiers} and \ref{equ: RoBERTa slsot filling} are a trainable weight matrix and a bias vector. RoBERTa is jointly fine-tuned via minimizing the sum of cross-entropy loss of domain, slot classifier and slot filling task, as shown in Equation \ref{equ: joint RoBERTa loss function}.
\begin{equation}
\label{equ: joint RoBERTa loss function}
\begin{aligned}
    & \emph{l}_{\textit{joint RoBERTa}} = \\
    & \sum_{M} (|| \hat{\emph{y}}^{d} - \emph{y}^{d} ||^{2} + || \hat{\emph{y}}^{s} - \emph{y}^{s} ||^{2} + \sum_{n=0}^{N} || \hat{\mathbf{y}}_{n}^{\textit{sf}} - \mathbf{y}^{\textit{sf}} ||^{2})
\end{aligned}
\end{equation}

%%%%%%%%%%%%%%%%%%%%%%%%%%%%%
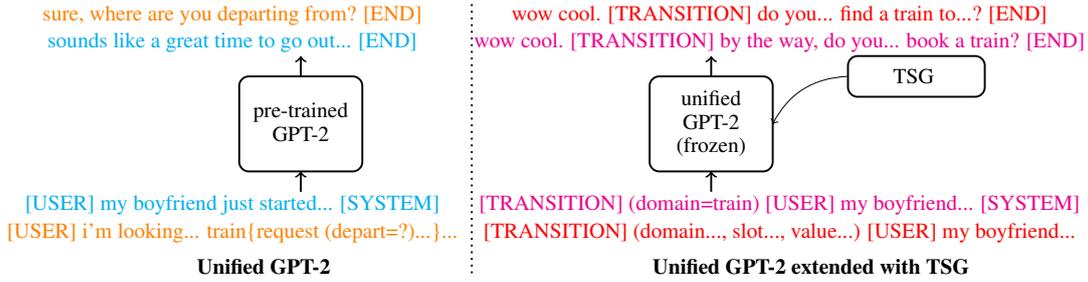
\begin{figure*}[h!!]
\centering
\footnotesize
\scalebox{0.9}{
\begin{tikzpicture}

\draw[->,thick] (0.5, 0.7) -- (0.5, 1.0);
\node at (-0.5, 1.2)[align=center]  {\textcolor{cyan}{sounds like a great time to go out... [END]}};
\node at (-0.5, 1.6)[align=center]  {\textcolor{orange}{sure, where are you departing from? [END]}};

\draw[thick,rounded corners]   (-0.4, -0.7) rectangle (1.4, 0.7);
\node at (0.5, 0.0)[align=center]  {\makecell[c]{pre-trained \\ GPT-2}};

\draw[->,thick] (0.5, -1.0) -- (0.5, -0.7);
\node at (-0.5, -1.2)[align=center]  {\textcolor{cyan}{[USER] my boyfriend just started... [SYSTEM]}};
\node at (-0.5, -1.6)[align=center]  {\textcolor{orange}{[USER] i'm looking... train\{request (depart=?)...\}...}};

\node at (0.0, -2.1)[align=center] { \textbf{Unified GPT-2} };

\draw[dotted,thick,black] (3.0, -2.2) -- (3.0, 1.8);

\draw[->,thick] (6.5, 0.7) -- (6.5, 1.0);
\node at (7.5, 1.2)[align=center]  {\textcolor{magenta}{wow cool. [TRANSITION] by the way, do you... book a train? [END]}};
\node at (7.5, 1.6)[align=center]  {\textcolor{red}{wow cool. [TRANSITION] do you... find a train to...? [END]}};

\draw[thick,rounded corners]   (5.6, -0.7) rectangle (7.4, 0.7);
\node at (6.5, 0.0)[align=center]  {\makecell[c]{unified \\ GPT-2 \\ (frozen)}};

\draw[thick,rounded corners]   (8.5, 0.4) rectangle (10.5, 1.0);
\node at (9.5, 0.7)[align=center]  { {TSG} };

\path[->] (8.5, 0.7) edge [bend right]  (7.4, 0.0);

\draw[->,thick] (6.5, -1.0) -- (6.5, -0.7);
\node at (7.5, -1.2)[align=center]  {\textcolor{magenta}{[TRANSITION] (domain=train) [USER] my boyfriend... [SYSTEM]}};
\node at (7.5, -1.6)[align=center]  {\textcolor{red}{[TRANSITION] (domain..., slot..., value...) [USER] my boyfriend...}};

\node at (8.0, -2.1)[align=center] { \textbf{Unified GPT-2 extended with TSG} };

\end{tikzpicture}
}
\caption{Architecture of unified GPT-2 and extended version with integrated TSG via Adapter tuning and transition prompt.
In the unified GPT-2, the \textcolor{orange}{orange} and \textcolor{cyan}{blue} represents the task-oriented and chit-chat example respectively. 
%The dialogue action (train\{request (depart=?)...\}) is also attached in the input for task-oriented response generation. 
Two transition scenarios for each dialogue are used as training data for the TS Adapter tuning, one is only transition domain (\textcolor{magenta}{magenta}) as prompt, the other is transition domain-slot-value (\textcolor{red}{red}) as prompt.
}
\label{fig: the architecture of unified GPT-2 and the extended version with transition sentence Adapter layers.}
\end{figure*}

\subsection{Conditional Random Fields (CRF)}
\label{subsec: CRF}

Beyond joint RoBERTa training for the transition domain/slot classification and slot filling tasks, we also use Conditional Random Fields (CRF) \citep{lafferty2001conditional}, to model the slot filling sequence jointly instead of decoding each slot filling label independently. CRF has been successfully used to exploit the dependencies within sequence labels corresponding to surrounding words and can highly improve the performance of slot filling task \citep{ma2016end}. 
In this work, the dropout layer is applied before feeding RoBERTa outputs into CRF layer. The Viterbi algorithm is used for decoding.

We only utilize the preceding chit-chat part of Prepended FusedChat for joint RoBERTa training. To better analyse the proposed TIE model, three different TIE models are trained. As shown in Table \ref{tab:The performance of transition domain classification, slot classification and slot filling in the NLU model.}, ``RoBERTa w/o slot filling'' only includes transition domain and slot classifiers; ``joint RoBERTa'' is jointly trained with domain, slot classifier and slot filling task together; and finally ``joint RoBERTa + CRF'' is our proposed final model, where the CRF is used for the slot filling task. All models are trained with two GPUs, the learning rate is $5\mathrm{e}{-5}$ and batch size is $32$. The best model of RoBERTa w/o slot filling is saved at epoch $5$ with early stopping. The joint RoBERTa is saved at epoch $4$ and joint RoBERTa + CRF at epoch $3$.

\section{Unified NLG extended with Transition Sentence Generator (TSG)}
\label{sec: unified GPT2 and transition sentence adapter}

This section firstly introduces the unified NLG model that can reply to both chit-chat and task-oriented requests. Then we mainly elaborate on the TSG integrated in unified NLG through efficient Adapter tuning and transition prompt technologies.
The extended NLG with TSG can generate a transition sentence given the transition information extracted by TIE to enable the system-initiated transition.
%When the task-related transition information is successfully extracted through TIE model, the transition information is converted into a transition prompt and fed into the unified NLG, meanwhile the TSG in the unified NLG is activated to generate a transition sentence and proactively guide this transition.
The details of unified NLG and the extension with TSG are shown in Figure \ref{fig: the architecture of unified GPT-2 and the extended version with transition sentence Adapter layers.}.

\subsection{Unified NLG}
\label{subsec: unified GPT-2}

We briefly presents the unified NLG model. By leveraging the entire FusedChat dataset \citep{young2022fusing}, where every dialogue includes interdependent chit-chat and task-oriented interaction, we tackle the unified generation problem through fine-tuning conditional GPT-2 \citep{radford2019language}. Given the FusedChat dataset $\mathcal{D^\prime} = \{(\emph{u}_{g}, \emph{d}_{g}, \emph{r}_{g})_{g=1}^{G}, (\emph{u}_{l}, \emph{r}_{l})_{l=1}^{L} \} $ with $G$ task-oriented samples (\textcolor{orange}{orange} in Figure \ref{fig: the architecture of unified GPT-2 and the extended version with transition sentence Adapter layers.}) and $L$ chit-chat samples (\textcolor{cyan}{blue} in Figure \ref{fig: the architecture of unified GPT-2 and the extended version with transition sentence Adapter layers.}), the goal is to build a unified model parameterized by $\theta$ to be able to respond to both chit-chat and task-oriented requests,
%shown in Equation \ref{equ: GPT-2 for CC and TO generation},
\begin{equation}
\label{equ: GPT-2 for CC and TO generation}
    p_{\theta}(\emph{r}) = 
    \begin{cases}
      \prod_{t=1}^{T} p_{\theta}(r_{t}|r_{<t}, \emph{u}, \emph{d}) & \text{if task-oriented} \\
      \prod_{t=1}^{T} p_{\theta}(r_{t}|r_{<t}, \emph{u}) & \text{if chit-chat}
    \end{cases}
\end{equation}
where $r_{<t}$ indicates all tokens before $t$. The $\emph{u}$ represents the dialogue context; $\emph{d}$ means the dialogue actions only exist in task-oriented data and $\emph{r}$ is the system response which includes $(r_{1}, ... r_{t}, ...)$ tokens with length $T$.

During the unified GPT-2 fine-tuning, we add $\text{[USER]}$ and $\text{[SYSTEM]}$ to the GPT-2 tokenizer to distinguish user utterances from system responses. At most three preceding dialogue turns are used as the dialogue context for response generation because of memory constraints. During training, the learning rate is $5\mathrm{e}{-5}$, batch size is $20$. The best model is saved at epoch $6$ with early stopping. We mix top-K sampling and top-p (nucleus) sampling \cite{holtzman2019curious} for decoding. We apply top-K of $5$ and top-p of $0.9$ for chit-chat response generation and top-K of $10$ and top-p of $0.5$ for task-oriented response generation respectively.

\subsection{Transition Prompt and Adapter Tuning}
\label{subsec: transition prompt and adapter tuning}

To enable the proactive capabilities, we integrate the efficient Adapter layers \citep{houlsby2019parameter, pfeiffer2021adapterfusion} into the unified GPT-2.
Adapter tuning freezes the parameters of a pre-trained model and injects lightweight modules between layers \citep{le2021lightweight} to enable a new capability.
Hence, the original performance of unified NLG for generating normal responses is retained without any loss. Meanwhile, the capability of generating transition sentences is extended through activating the newly added Adapter layers.
To further explicitly control the transition sentence generation, the prompt learning \cite{liu2021pre, li2022personalized} is used.
More precisely, when the TIE model successfully detects the user intention requiring a task-related service, the integrated Adapter layers are activated meanwhile the transition information extracted via TIE
is converted into prompt input to generate a transition sentence to proactively establish the transition from chit-chat to task-oriented.

\subsubsection{Transition Prompt}
\label{suubsubsec: transition prompt}

%To enable the transition sentence generation at the transition turn, we need to firstly convert the output of the NLU model (Section \ref{sec: NLU+CRF}) to the transition prompt, then prepend it to the dialogue context as input to feed into the unified GPT-2. This section will briefly introduce what the transition prompt is.

Prompt learning can efficiently adapt a given task to pre-trained models without modifying the structure of models \cite{lester2021power}.
In this work, we only convert the task-related transition information extracted by TIE to the transition prompt, which is a part of the input for the generation model that explicitly guides the transition sentence generation.

We add a special token $\text{[TRANSITION]}$ into the GPT-2 tokenizer and insert this token into the first place of the task-related transition prompt. Two different types of transition prompt are as follows:
%\footnote{Two transition cases are proposed here just in case when the predicted domain-slot-value combination does not make sense, we can have another alternative for generating a domain guided transition sentence.}
\begin{enumerate}
    \item
    When only the transition domain information is available, the prompt is like ``[TRANSITION] ( domain = train ) '', where ``train'' is the extracted transition domain (\textcolor{magenta}{magenta} input in Figure \ref{fig: the architecture of unified GPT-2 and the extended version with transition sentence Adapter layers.}).
    
    \item
    When transition domain, slot and value are all extracted via TIE model, then the prompt is like ``[TRANSITION] ( domain = train,  slot = destination, value = Norwich ) '', where the transition domain is ``train'', slot is ``destination'' along with the value ``Norwich'' (\textcolor{red}{red} input in Figure \ref{fig: the architecture of unified GPT-2 and the extended version with transition sentence Adapter layers.}).
\end{enumerate}
The dialogue context is prepended with the transition prompt to be the input of the generation model. In addition, $\text{[TRANSITION]}$ is also used to separate the transition sentence from normal response at transition turn (responses of \textcolor{magenta}{magenta} and \textcolor{red}{red} examples in Figure \ref{fig: the architecture of unified GPT-2 and the extended version with transition sentence Adapter layers.}). Hence, the $\text{[TRANSITION]}$ in prompt inputs is a signal for the generation model that it is a good moment to guide the transition to task-oriented service because TIE extracts task-related information, while the $\text{[TRANSITION]}$ in generated responses is a signal to demonstrate that the NLG model is able to generate a transition sentence for proactive transition.

%that current is a good moment to guide the transition because TIE extracts valuable task-related information, meanwhile to generate a reasonable transition sentence besides the normal response through triggering Adapter layers to proactively guide this transition.

\subsubsection{TSG through Adapter Tuning}
\label{subsubsec: transition adapter tuning}

We utilize the AdapterHub \citep{pfeiffer2020adapterhub}, which is a framework that can easily integrate Adapters into pre-trained Transformer-based models \citep{vaswani2017attention}. The Houlsby Adapter \citep{houlsby2019parameter} includes two bottleneck adapters in each transformer layer, one after the multi-head attention sub-layer and the other after the feed-forward sub-layer. The Pfeiffer Adapter \citep{pfeiffer2021adapterfusion} only includes the adapter after the feed-forward sub-layer. Only 1\% (Pfeiffer) and 2\% (Houlsby) parameters are updated during Adapter tuning with frozen unified GPT-2. Hence, we can efficiently integrate the transition sentence generation into the unified GPT-2, while keeping the original capabilities of generating normal chit-chat and task-oriented responses by deactivating the Adapter layers.

Only the generation at the transition turn is utilized for the training of TSG. Every dialogue has two transition cases: One only consists of transition domain as prompt (\textcolor{magenta}{magenta} input in Figure \ref{fig: the architecture of unified GPT-2 and the extended version with transition sentence Adapter layers.}) and the other consists of transition domain-slot-value as prompt (\textcolor{red}{red} input in Figure \ref{fig: the architecture of unified GPT-2 and the extended version with transition sentence Adapter layers.}). We prepend the transition prompt before the preceding chit-chat context as input. The response includes a normal chit-chat response as well as a transition sentence separated with $\text{[TRANSITION]}$ (the response of \textcolor{red}{red} and \textcolor{magenta}{magenta} examples in Figure \ref{fig: the architecture of unified GPT-2 and the extended version with transition sentence Adapter layers.}). For the TSG, the Houlsby and Pfeiffer Adapters are both trained with the learning rate $5\mathrm{e}{-5}$, batch size $20$. The best models are both saved at epoch $16$ (early stopping). We apply top-K of $5$ and top-p of $0.9$ for the response generation at the transition turn.

\begin{table*}
\centering
\footnotesize
\scalebox{0.9}{
\begin{tabular}{cccccccc}
\toprule
    
    & \multicolumn{2}{c}{domain classifier} & \multicolumn{2}{c}{slot classifier } & \multicolumn{2}{c}{slot filling}  & \\
    \cmidrule(lr{.75em}){2-3}  \cmidrule(lr{.75em}){4-5} \cmidrule(lr{.75em}){6-7} 

    & accuracy & weighted f1 & accuracy & weighted f1 & sen\_sf\_acc & sf\_f1  & semantic\_acc\\
         
\midrule
    
    RoBERTa w/o slot filling & 78.57\% & 79.57\% & 66.52\% & 66.84\% & -- & -- & -- \\
         
    joint RoBERTa & 82.41\% & 82.92\% & 71.86\% & 73.84\% & 68.02\% & 48.64\%  & 61.94\% \\
         
    joint RoBERTa + CRF & \textbf{93.71\%} & \textbf{94.15\%} & \textbf{82.41\%} & \textbf{82.30\%} & \textbf{80.28\%} & \textbf{61.82\%} & \textbf{73.67\%} \\
         
\bottomrule
\end{tabular}}
\caption{Performance of transition domain/slot classification and slot filling task in different TIE models.}
% The first one is that RoBERTa is only fine-tuned for domain and slot classification without slot filling task; the second one is that RoBERTa is jointly fine-tuned for slot filling task and domain/slot classification; Beyond that, the CRF technology is utilized in the third TIE model.
\label{tab:The performance of transition domain classification, slot classification and slot filling in the NLU model.}
\end{table*}

\section{Results Comparison}
\label{sec: results comparison}

This section evaluates this work and provides detailed performance comparison from different perspectives. We firstly evaluate different TIE models and different generation models separately with test Prepended FusedChat. Then we further evaluate the combined performance of the best TIE model and generation model only at transition turns.

\begin{table*}[]
\centering
\footnotesize
\newcolumntype{a}{>{\columncolor[gray]{0.9}}c}
\scalebox{0.75}
{
\begin{tabular}{ll ccccc ccacca}
\toprule
    & & \multicolumn{2}{c}{Chit-Chat} & \multicolumn{3}{c}{Task-Oriented} & \multicolumn{3}{c}{domain TS} & \multicolumn{3}{c}{domain-slot-value TS} \\
    \cmidrule(lr{.75em}){3-4}  \cmidrule(lr{.75em}){5-7} \cmidrule(lr{.75em}){8-10}  \cmidrule(lr{.75em}){11-13}
    
    & & \makecell{Dis-1} &
    \makecell{Dis-2} &
    \makecell{BLUE} &
    \makecell{Meteor} &
    \makecell[c]{BERTScore \\ (F1)} &
    \makecell[c]{BERTScore \\ (F1)} &
    \makecell[c]{\emph{transition} \\ \emph{accuracy}} & 
    \makecell[c]{\emph{d} \\ \emph{accuracy}} &
    \makecell[c]{BERTScore \\ (F1)} &
    \makecell[c]{\emph{transition} \\ \emph{accuracy}} &
    \makecell[c]{\emph{d-v} \\ \emph{accuracy}} \\
    
\midrule
    
    \multicolumn{2}{c}{unified GPT-2} & \textbf{1.74\%} &  \textbf{12.70\%} & \textbf{34.77\%} &  \textbf{55.65\%} &  \textbf{93.20\%} & -- & -- & -- & -- & -- & -- \\

%\midrule
    
    \multirow{2}{*}{retrain} &
    w/o TP  &  1.67\%  &  11.41\%  &  32.86\%  &  53.52\%  & 92.91\%  & 88.82\% & 98.25\% & \cellcolor[gray]{0.9}58.19\% & 89.29\% & 98.97\% & \cellcolor[gray]{0.9}30.15\% \\
    & w/ TP &  1.60\%  &  11.18\%  &  32.58\%  &  53.33\%  & 92.94\%  & 90.19\% & 98.43\% & \cellcolor[gray]{0.9}99.21\% & 91.70\% & 98.79\% & \cellcolor[gray]{0.9}92.63\% \\

\midrule
    
    \multirow{2}{*}{\makecell[c]{TSG \\ (Houlsby)}} &
    w/o TP & \textbf{1.74\%} &  \textbf{12.70\%} & \textbf{34.77\%} &  \textbf{55.65\%} &  \textbf{93.20\%} & 89.04\% & 98.67\% & \cellcolor[gray]{0.9}62.48\%  & 89.40\%  & \textbf{99.34\%}  & \cellcolor[gray]{0.9}27.19\% \\
    
    & w/ TP & \textbf{1.74\%} &  \textbf{12.70\%} & \textbf{34.77\%} &  \textbf{55.65\%} &  \textbf{93.20\%} & 90.28\%  & \textbf{99.40\%} & \cellcolor[gray]{0.9}99.15\%  & \textbf{91.84\%}  & 99.21\%  & \cellcolor[gray]{0.9}\textbf{96.80\%} \\

\midrule
    
    \multirow{2}{*}{\makecell[c]{TSG \\ (Pfeiffer)}} &
    w/o TP & \textbf{1.74\%} &  \textbf{12.70\%} & \textbf{34.77\%} &  \textbf{55.65\%} &  \textbf{93.20\%} & 88.90\%  &  97.82\%  &  \cellcolor[gray]{0.9} 59.52\% & 89.33\% & 98.25\% & \cellcolor[gray]{0.9}25.98\% \\
    
    & w/ TP & \textbf{1.74\%} &  \textbf{12.70\%} & \textbf{34.77\%} &  \textbf{55.65\%} &  \textbf{93.20\%} & \textbf{90.34\%}  & 98.13\%  &  \cellcolor[gray]{0.9} \textbf{99.70\%} & 91.83\% & 98.43\% & \cellcolor[gray]{0.9}96.62\% \\
    
\bottomrule
\end{tabular}
}
\caption{Performance of different NLG models, including unified GPT-2 and retrained without Adapter, extended with Houslby and Pfeiffer TSG separately, and all with transition prompt (w/ TP) and w/o TP respectively.
%The unified GPT-2; retrained version without Adapter, along with transition prompt (retrain w/ TP) and without transition prompt (retrain w/o TP) respectively; and our proposed extended version with TSG, the unified GPT-2 integrated with Adapter (Houslby and Pfeiffer separately), along with transition prompt (w/ TP) and without transition prompt (w/o TP) respectively.
}
\label{tab: performance comparison of unified GPT-2 with different adapter.}
\end{table*}

\subsection{TIE models}
\label{subsec: performance of NLU model}

Table \ref{tab:The performance of transition domain classification, slot classification and slot filling in the NLU model.} shows the performance comparison of different TIE models. We use classification accuracy and weighted F1 score to evaluate the performance of transition domain and slot classifiers. Slot filling F1 (sf\_f1) score is widely used to evaluate the slot filling task \citep{chen2019bert}. In addition, we also use sentence-level slot filling accuracy (sen\_sf\_acc), which is the ratio of the number of dialogues correctly labelled slot filling to the total number of dialogues. The overall performance of the TIE model is evaluated using sentence-level semantic accuracy (semantic\_acc) \citep{yu2010sequential, weld2021survey} which measures the proportion of the correctly predicted triples of transition domain, slot, and extracted slot filling values (including ``O'' labels). %transition domain and slots which are jointly predicted correctly, meanwhile all the slot filling (including ``O'' labels) in a dialogue are correctly predicted.

The performance comparison in Table \ref{tab:The performance of transition domain classification, slot classification and slot filling in the NLU model.} demonstrates that joint RoBERTa with CRF as the TIE model achieves the best performance over transition domain classifier, slot classifier and slot filling task. It is surprising that not only the slot filling task benefits from the CRF. The performance of transition domain and slot classifiers is improved in the multi-task learning as well. 
%at the word level

\begin{table*}
\centering
\footnotesize
\scalebox{0.8}
{\begin{tabular}{cc cccc cccccc}

\toprule
    & & \multicolumn{4}{c}{TIE (joint RoBERTa + CRF)} & \multicolumn{6}{c}{Extended GPT-2 with TSG} \\
    \cmidrule(lr{.75em}){3-6}  \cmidrule(lr{.75em}){7-12} 

    & & \multicolumn{1}{c}{domain cls} & \multicolumn{1}{c}{slot cls} 
    & \multicolumn{1}{c}{slot filling} & \multicolumn{1}{c}{}
    & \multicolumn{3}{c}{domain TS} 
    & \multicolumn{3}{c}{domain-slot-value TS}\\
    \cmidrule(lr{.75em}){3-3} \cmidrule(lr{.75em}){4-4}  \cmidrule(lr{.75em}){5-5} 
    \cmidrule(lr{.75em}){7-9} 
    \cmidrule(lr{.75em}){10-12}

    & & \makecell[c]{accuracy}
    & \makecell[c]{accuracy} &
    %\makecell[c]{TV (idx) \\ accuracy} &
    %\makecell[c]{TV (idx\&sf) \\ accuracy} & 
    \makecell[c]{sf\_f1} &
    \makecell[c]{semantic\_acc} & 
    
    \makecell[c]{BERTScore \\ (F1)} &
    \makecell[c]{\emph{transition} \\ \emph{accuracy}} & \makecell[c]{\emph{d} \\ \emph{accuracy}} &
    \makecell[c]{BERTScore \\ (F1)} &
    \makecell[c]{\emph{transition} \\ \emph{accuracy}} & \makecell[c]{\emph{d-v} \\ \emph{accuracy}} \\
    
\midrule

    \makecell[c]{TSG \\ (Houlsby)}  & {w/ TP}
    & \multirow{4}{*}{93.35\%} &
    \multirow{4}{*}{65.56\%} &
    \multirow{4}{*}{64.71\%} &
    \multirow{4}{*}{50.15\%} &
    %\multirow{4}{*}{91.54\%} & ## TV (idx) \\ accuracy
    %\multirow{4}{*}{62.84\%} & ## TV (idx\&sf) \\ accuracy
    90.10\%  & 99.40\%  & 92.87\%
    & 91.10\% & 99.21\% & 82.78\% \\

    \makecell[c]{TSG \\ (Pfeiffer)} & {w/ TP} 
    & & & & &
    90.08\%  & 98.49\%  & 93.53\%
    & 91.25\% & 98.37\% & 83.02\% \\

\bottomrule
\end{tabular}}
\caption{Overall performance of combined TIE and extended GPT-2 with TSG at transition turns.
%The TIE predicts transition domain (TD), transition slot (TS) and extracts transition values (TV) from dialogue history. These generated transition information by the TIE model as transition prompt to guide transition sentence generation in the unified GPT-2 integrated with Houslby and Pfeiffer Adapter respectively.
}
\label{tab: performance comparison of combined TIE and GPT-2 model.}
\end{table*}

\subsection{Generation models}
\label{subsec: performance of NLG model}

To evaluate generated chit-chat responses, Distinct-1 (Dis-1) and Distinct-2 (Dis-2) \citep{li2016diversity} are used to measure the proportion of the distinct unigrams and bigrams in all the generated results to indicate diversity. To evaluate generated task-oriented responses, two $\emph{N}$-gram matching metrics, BLEU \citep{papineni2002bleu} and Meteor \citep{banerjee2005meteor} are used to evaluate the overall quality of task-oriented generations. In addition, a machine learned automatic metric, BERTScore \citep{zhang2019bertscore} is also utilized to evaluate task-oriented and transition sentence generations.

Beyond that, we propose several automatic metrics to evaluate transition sentence generations. \textit{Transition accuracy} detects whether the generated response at transition turn includes the $\text{[TRANSITION]}$ special token. With $\text{[TRANSITION]}$, we can split the transition sentence from the normal response. This metric can measure high-level capability if the model can generate a transition sentence to proactively switch to a task-oriented service. \textit{d accuracy} detects if the domain guided transition sentence includes the specific domain keyword. \textit{d-v accuracy} detects if the transition domain-slot-value guided transition sentence includes specific domain and value keywords both. \textit{d accuracy} and \textit{d-v accuracy} can evaluate the capability of the proposed transition prompt for explicitly controlling transition sentence generation to a large extent. 

We found that almost all generated transition sentences by TSG with TP are of high quality and include the extracted transition information (several cases are shown in Table \ref{tab: use cases study to show the ability to generate transition sentences to enable the system-initiated transitions from chit-chat to task-oriented services.} in Appendix), instead of generic transition responses like ``Do you need anything else?'' or ``Do you need some help?''.
%because the extracted transition information should appear in the generated transition sentence to explicitly guide the switching to task-oriented service rather than a generic transition response like ``Do you need anything else?'' or ``Do you need some help?''.

To better understand the performance of our models, % our proposed TSG extended in unified GPT-2 via Adapter tuning, 
we also retrain the unified GPT-2 without Adapter to enable its transition sentence generation (without TSG). From the comparison between the retrained model and unified GPT-2 in Table \ref{tab: performance comparison of unified GPT-2 with different adapter.}, we can see that the performance on chit-chat and task-oriented response generations has a loss, even though the retrained GPT-2 is still able to generate transition sentences. In contrast, our TSG extended in unified GPT-2 through Adapter tuning can retain the original capability for chit-chat and task-oriented generations, while maintaining a better performance on transition sentence generation.% compared to the retrained models. 
In addition, the retraining is not memory-efficient, while TSG only updates the Adapter parameters with frozen GPT-2.

To better assess the effects of our proposed transition prompt method, we retrain the model and extend TSG both along with the transition prompt (w/ TP) and without the transition prompt (w/o TP) respectively. Through the comparison between w/o TP and w/ TP in different models (highlighted in gray background in Table \ref{tab: performance comparison of unified GPT-2 with different adapter.}), the \textit{d accuracy} and \textit{d-v accuracy} metrics are highly improved with transition prompt guidance. This demonstrates that transition prompt can explicitly control the transition sentence generation. The performance comparison between Pfeiffer and Houlsby Adapter tuning has no big difference, however, the Pfeiffer Adapter uses only half of the trainable parameters, and is therefore the more effective choice for this work. 
%The generation of domain guided TS and domain-slot-value guided TS in the TSG (w/ TP) both have similar and promising performance. This provides us an alternative when the extracted domain-slot-value combination does not make sense, we can only generate a domain guided TS.

\subsection{Combined TIE and generation model}
\label{subsec: Performance of combined model}

To better reflect the overall performance of this work, we evaluate the combined TIE and generation models at transition turns, i.e., given the preceding chit-chat, the TIE model predicts transition domain/slot and extracts values, then this generated transition information by TIE is used as the transition prompt to guide transition sentence generation at the transition turn. Table \ref{tab: performance comparison of combined TIE and GPT-2 model.} shows the combined performance of TIE and unified GPT-2 with Houlsby and Pfeiffer TSG, respectively. 
%Domain and slot accuracy evaluate how many dialogues in test data are correctly predicted transition domain and slot respectively. Two different kinds of value accuracy are shown in Table \ref{tab: performance comparison of combined TIE and GPT-2 model.}. The value (idx) accuracy indicates the proportion of test dialogues, where the positions of transition values are correctly predicted regardless of whether the slot filling labels are correctly predicted. While the value (idx\&sf) accuracy takes both positions and slot filling labels of transition values into account. 

Given the higher domain accuracy compared to slot accuracy, it is sensible to only use domain prediction as transition information to guide transition sentence generation when generated transition slot or extracted values are not reliable. This also validates our initial idea to propose two kinds of transition prompts.
Regarding the lower slot accuracy, we found that the TIE model tends to confuse 
 ``destination'' and ``departure'' under the ``train'' domain; over 60\% of slot misjudged dialogues are in these cases.
This would further affect the overall performance of the TIE model, which is shown by the semantic\_acc metric.

Each Prepended FusedChat dialogue has only one turn for the transition from chit-chat to task-oriented. We directly define this turn as the transition turn, where the initiative dialogue model proactively switches to a task-oriented service through generating a transition sentence. %Hence, we have not seriously discussed WHEN to make the system-initiated transitions is appropriate.
%This would be addressed in the future work.
%However, 
Also, dialogue interactions could be more sophisticated in real life and it is difficult to accurately define the most appropriate moment to initiate a proactive transition.
Furthermore, it gets more complicated if there are multiple transitions in one dialogue.
A further, deeper investigation of appropriate moments for a dialogue mode transition will be done in future work.
%This discussion can even be extended to if the customer is happy with the system-initiated transitions even if dialogue systems successfully anticipate in user intentions.
%To sum up, the topic of initiative transitions in dialogue systems is promising but also challenging.

\section{Conclusion}
\label{sec: conclusion}

This work investigates the dialogue transition from chit-chat to task-oriented initiated by a dialogue agent. We build a TIE model adapted from pre-trained RoBERTa to keep track of the preceding chit-chat and predict transition domain, slot, while extracting the specific value from the chit-chat history via slot filling task. A unified generation model adapted from the pre-trained GPT-2 is built and extended its proactive capability for transition sentence generation through efficient Adapter tuning and transition prompt learning. Our proposed work shows promising performance both on transition information extraction and transition sentence generation. We will continue working on system-initiated transitions in other dialogue scenarios in the future.

\section{Ethics Statement}
\label{sec: Ethics Statement}
This work develops proactive transitions from chit-chat to task-oriented dialogue in a unified dialogue system. Proactivity is always desired during the development of voice assistants. It can improve user interactive experience and serve users more efficiently. The dataset used in this work is public available and manually collected. Furthermore, our research is limited to a specific case, i.e, the user starts casual chat and eventually switches to a task-oriented service. However, more hidden challenges and ethics issues should be discussed further in the real scenarios. Would users prefer to be proactively served if the dialogue system successfully detects the user intention? Will they feel their privacy is violated if the dialogue system proactively provides task-related services? Such potential issues could be addressed by asking for user consent before providing the proactive interaction, which raises the additional question how many users would turn on such a feature from the start. %These potential issues can be easily addressed by asking users, do they agree to turn on the proactive feature at the very beginning? Only users with this approval can be served proactively.

%\subsection{Appendices}
%\section*{Acknowledgements}

\bibliography{anthology,custom}

\begin{thebibliography}{43}
\expandafter\ifx\csname natexlab\endcsname\relax\def\natexlab#1{#1}\fi

\bibitem[{Banerjee and Lavie(2005)}]{banerjee2005meteor}
Satanjeev Banerjee and Alon Lavie. 2005.
\newblock Meteor: An automatic metric for mt evaluation with improved
  correlation with human judgments.
\newblock In \emph{Proceedings of the acl workshop on intrinsic and extrinsic
  evaluation measures for machine translation and/or summarization}, pages
  65--72.

\bibitem[{Budzianowski et~al.(2018)Budzianowski, Wen, Tseng, Casanueva, Ultes,
  Ramadan, and Gasic}]{budzianowski2018multiwoz}
Pawe{\l} Budzianowski, Tsung-Hsien Wen, Bo-Hsiang Tseng, I{\~n}igo Casanueva,
  Stefan Ultes, Osman Ramadan, and Milica Gasic. 2018.
\newblock Multiwoz-a large-scale multi-domain wizard-of-oz dataset for
  task-oriented dialogue modelling.
\newblock In \emph{Proceedings of the 2018 Conference on Empirical Methods in
  Natural Language Processing}, pages 5016--5026.

\bibitem[{Chen et~al.(2019)Chen, Zhuo, and Wang}]{chen2019bert}
Qian Chen, Zhu Zhuo, and Wen Wang. 2019.
\newblock Bert for joint intent classification and slot filling.
\newblock \emph{arXiv preprint arXiv:1902.10909}.

\bibitem[{Chiu et~al.(2022)Chiu, Li, Lin, and Chen}]{chiu2022salesbot}
Ssu Chiu, Maolin Li, Yen-Ting Lin, and Yun-Nung Chen. 2022.
\newblock Salesbot: Transitioning from chit-chat to task-oriented dialogues.
\newblock In \emph{Proceedings of the 60th Annual Meeting of the Association
  for Computational Linguistics (Volume 1: Long Papers)}, pages 6143--6158.

\bibitem[{De~Mori et~al.(2008)De~Mori, Bechet, Hakkani-Tur, McTear, Riccardi,
  and Tur}]{de2008spoken}
Renato De~Mori, Fr{\'e}d{\'e}ric Bechet, Dilek Hakkani-Tur, Michael McTear,
  Giuseppe Riccardi, and Gokhan Tur. 2008.
\newblock Spoken language understanding.
\newblock \emph{IEEE Signal Processing Magazine}, 25(3):50--58.

\bibitem[{Goo et~al.(2018)Goo, Gao, Hsu, Huo, Chen, Hsu, and
  Chen}]{goo2018slot}
Chih-Wen Goo, Guang Gao, Yun-Kai Hsu, Chih-Li Huo, Tsung-Chieh Chen, Keng-Wei
  Hsu, and Yun-Nung Chen. 2018.
\newblock Slot-gated modeling for joint slot filling and intent prediction.
\newblock In \emph{Proceedings of the 2018 Conference of the North American
  Chapter of the Association for Computational Linguistics: Human Language
  Technologies, Volume 2 (Short Papers)}, pages 753--757.

\bibitem[{Holtzman et~al.(2019)Holtzman, Buys, Du, Forbes, and
  Choi}]{holtzman2019curious}
Ari Holtzman, Jan Buys, Li~Du, Maxwell Forbes, and Yejin Choi. 2019.
\newblock The curious case of neural text degeneration.
\newblock In \emph{International Conference on Learning Representations}.

\bibitem[{Houlsby et~al.(2019)Houlsby, Giurgiu, Jastrzebski, Morrone,
  De~Laroussilhe, Gesmundo, Attariyan, and Gelly}]{houlsby2019parameter}
Neil Houlsby, Andrei Giurgiu, Stanislaw Jastrzebski, Bruna Morrone, Quentin
  De~Laroussilhe, Andrea Gesmundo, Mona Attariyan, and Sylvain Gelly. 2019.
\newblock Parameter-efficient transfer learning for nlp.
\newblock In \emph{International Conference on Machine Learning}, pages
  2790--2799. PMLR.

\bibitem[{Kenton and Toutanova(2019)}]{kenton2019bert}
Jacob Devlin Ming-Wei~Chang Kenton and Lee~Kristina Toutanova. 2019.
\newblock Bert: Pre-training of deep bidirectional transformers for language
  understanding.
\newblock In \emph{Proceedings of NAACL-HLT}, pages 4171--4186.

\bibitem[{Lafferty et~al.(2001)Lafferty, McCallum, and
  Pereira}]{lafferty2001conditional}
John Lafferty, Andrew McCallum, and Fernando~CN Pereira. 2001.
\newblock Conditional random fields: Probabilistic models for segmenting and
  labeling sequence data.

\bibitem[{Le et~al.(2021)Le, Pino, Wang, Gu, Schwab, and
  Besacier}]{le2021lightweight}
Hang Le, Juan Pino, Changhan Wang, Jiatao Gu, Didier Schwab, and Laurent
  Besacier. 2021.
\newblock Lightweight adapter tuning for multilingual speech translation.
\newblock In \emph{The Joint Conference of the 59th Annual Meeting of the
  Association for Computational Linguistics and the 11th International Joint
  Conference on Natural Language Processing (ACL-IJCNLP 2021)}.

\bibitem[{Lester et~al.(2021)Lester, Al-Rfou, and Constant}]{lester2021power}
Brian Lester, Rami Al-Rfou, and Noah Constant. 2021.
\newblock The power of scale for parameter-efficient prompt tuning.
\newblock In \emph{Proceedings of the 2021 Conference on Empirical Methods in
  Natural Language Processing}, pages 3045--3059.

\bibitem[{Li et~al.(2016)Li, Galley, Brockett, Gao, and
  Dolan}]{li2016diversity}
Jiwei Li, Michel Galley, Chris Brockett, Jianfeng Gao, and William~B Dolan.
  2016.
\newblock A diversity-promoting objective function for neural conversation
  models.
\newblock In \emph{Proceedings of the 2016 Conference of the North American
  Chapter of the Association for Computational Linguistics: Human Language
  Technologies}, pages 110--119.

\bibitem[{Li et~al.(2022)Li, Zhang, and Chen}]{li2022personalized}
Lei Li, Yongfeng Zhang, and Li~Chen. 2022.
\newblock Personalized prompt learning for explainable recommendation.
\newblock \emph{arXiv preprint arXiv:2202.07371}.

\bibitem[{Lin et~al.(2021)Lin, Madotto, Bang, and Fung}]{lin2021adapter}
Zhaojiang Lin, Andrea Madotto, Yejin Bang, and Pascale Fung. 2021.
\newblock The adapter-bot: All-in-one controllable conversational model.
\newblock In \emph{Proceedings of the AAAI Conference on Artificial
  Intelligence}, volume~35, pages 16081--16083.

\bibitem[{Liu and Lane(2016)}]{liu2016attention}
Bing Liu and Ian Lane. 2016.
\newblock Attention-based recurrent neural network models for joint intent
  detection and slot filling.
\newblock \emph{Interspeech 2016}, pages 685--689.

\bibitem[{Liu et~al.(2021)Liu, Yuan, Fu, Jiang, Hayashi, and
  Neubig}]{liu2021pre}
Pengfei Liu, Weizhe Yuan, Jinlan Fu, Zhengbao Jiang, Hiroaki Hayashi, and
  Graham Neubig. 2021.
\newblock Pre-train, prompt, and predict: A systematic survey of prompting
  methods in natural language processing.
\newblock \emph{arXiv preprint arXiv:2107.13586}.

\bibitem[{Liu et~al.(2023)Liu, Ultes, Minker, and Maier}]{liu2023unified}
Ye~Liu, Stefan Ultes, Wolfgang Minker, and Wolfgang Maier. 2023.
\newblock Unified conversational models with system-initiated transitions
  between chit-chat and task-oriented dialogues.
\newblock \emph{arXiv preprint arXiv:2307.01664}.

\bibitem[{Liu et~al.(2022)Liu, Yang, Maier, Minker, and
  Ultes}]{liu-etal-2022-system}
Ye~Liu, Yung-Ching Yang, Wolfgang Maier, Wolfgang Minker, and Stefan Ultes.
  2022.
\newblock \href {http://semdial.org/anthology/Z22-Liu_semdial_0034.pdf} {On
  system-initiated transitions in a unified natural language generation model
  for dialogue systems}.
\newblock In \emph{Proceedings of the 26th Workshop on the Semantics and
  Pragmatics of Dialogue - Poster Abstracts}, Dublin, Ireland. SEMDIAL.

\bibitem[{Liu et~al.(2019)Liu, Ott, Goyal, Du, Joshi, Chen, Levy, Lewis,
  Zettlemoyer, and Stoyanov}]{liu2019roberta}
Yinhan Liu, Myle Ott, Naman Goyal, Jingfei Du, Mandar Joshi, Danqi Chen, Omer
  Levy, Mike Lewis, Luke Zettlemoyer, and Veselin Stoyanov. 2019.
\newblock Roberta: A robustly optimized bert pretraining approach.
\newblock \emph{arXiv preprint arXiv:1907.11692}.

\bibitem[{Liu et~al.(2020)Liu, Wang, Niu, Wu, Che, and Liu}]{liu2020towards}
Zeming Liu, Haifeng Wang, Zheng-Yu Niu, Hua Wu, Wanxiang Che, and Ting Liu.
  2020.
\newblock Towards conversational recommendation over multi-type dialogs.
\newblock In \emph{Proceedings of the 58th Annual Meeting of the Association
  for Computational Linguistics}, pages 1036--1049.

\bibitem[{Ma and Hovy(2016)}]{ma2016end}
Xuezhe Ma and Eduard Hovy. 2016.
\newblock End-to-end sequence labeling via bi-directional lstm-cnns-crf.
\newblock In \emph{Proceedings of the 54th Annual Meeting of the Association
  for Computational Linguistics (Volume 1: Long Papers)}, pages 1064--1074.

\bibitem[{Mesnil et~al.(2014)Mesnil, Dauphin, Yao, Bengio, Deng, Hakkani-Tur,
  He, Heck, Tur, Yu et~al.}]{mesnil2014using}
Gr{\'e}goire Mesnil, Yann Dauphin, Kaisheng Yao, Yoshua Bengio, Li~Deng, Dilek
  Hakkani-Tur, Xiaodong He, Larry Heck, Gokhan Tur, Dong Yu, et~al. 2014.
\newblock Using recurrent neural networks for slot filling in spoken language
  understanding.
\newblock \emph{IEEE/ACM Transactions on Audio, Speech, and Language
  Processing}, 23(3):530--539.

\bibitem[{Nothdurft et~al.(2015)Nothdurft, Ultes, and Minker}]{2014fn02}
Florian Nothdurft, Stefan Ultes, and Wolfgang Minker. 2015.
\newblock \href {http://www.ep.liu.se/ecp/110/010/ecp15110010.pdf} {Finding
  appropriate interaction strategies for proactive dialogue systems---an open
  quest}.
\newblock In \emph{Proc. of the 2nd European and the 5th Nordic Symposium on
  Multimodal Communication 2014}, pages 73--80. LiU Electronic Press.

\bibitem[{Papineni et~al.(2002)Papineni, Roukos, Ward, and
  Zhu}]{papineni2002bleu}
Kishore Papineni, Salim Roukos, Todd Ward, and Wei-Jing Zhu. 2002.
\newblock Bleu: a method for automatic evaluation of machine translation.
\newblock In \emph{Proceedings of the 40th annual meeting of the Association
  for Computational Linguistics}, pages 311--318.

\bibitem[{Pfeiffer et~al.(2021)Pfeiffer, Kamath, R{\"u}ckl{\'e}, Cho, and
  Gurevych}]{pfeiffer2021adapterfusion}
Jonas Pfeiffer, Aishwarya Kamath, Andreas R{\"u}ckl{\'e}, Kyunghyun Cho, and
  Iryna Gurevych. 2021.
\newblock Adapterfusion: Non-destructive task composition for transfer
  learning.
\newblock In \emph{Proceedings of the 16th Conference of the European Chapter
  of the Association for Computational Linguistics: Main Volume}, pages
  487--503.

\bibitem[{Pfeiffer et~al.(2020)Pfeiffer, R{\"u}ckl{\'e}, Poth, Kamath,
  Vuli{\'c}, Ruder, Cho, and Gurevych}]{pfeiffer2020adapterhub}
Jonas Pfeiffer, Andreas R{\"u}ckl{\'e}, Clifton Poth, Aishwarya Kamath, Ivan
  Vuli{\'c}, Sebastian Ruder, Kyunghyun Cho, and Iryna Gurevych. 2020.
\newblock Adapterhub: A framework for adapting transformers.
\newblock In \emph{Proceedings of the 2020 Conference on Empirical Methods in
  Natural Language Processing: System Demonstrations}, pages 46--54.

\bibitem[{Radford et~al.(2019)Radford, Wu, Child, Luan, Amodei, and
  Sutskever}]{radford2019language}
Alec Radford, Jeffrey Wu, Rewon Child, David Luan, Dario Amodei, and Ilya
  Sutskever. 2019.
\newblock Language models are unsupervised multitask learners.
\newblock \emph{OpenAI blog}, 1(8):9.

\bibitem[{Ramshaw and Marcus(1999)}]{ramshaw1999text}
Lance~A Ramshaw and Mitchell~P Marcus. 1999.
\newblock Text chunking using transformation-based learning.
\newblock In \emph{Natural language processing using very large corpora}, pages
  157--176. Springer.

\bibitem[{Sha and Pereira(2003)}]{sha2003shallow}
Fei Sha and Fernando Pereira. 2003.
\newblock Shallow parsing with conditional random fields.
\newblock In \emph{Proceedings of the 2003 Human Language Technology Conference
  of the North American Chapter of the Association for Computational
  Linguistics}, pages 213--220.

\bibitem[{Shuster et~al.(2020)Shuster, Ju, Roller, Dinan, Boureau, and
  Weston}]{shuster2020dialogue}
Kurt Shuster, Da~Ju, Stephen Roller, Emily Dinan, Y-Lan Boureau, and Jason
  Weston. 2020.
\newblock The dialogue dodecathlon: Open-domain knowledge and image grounded
  conversational agents.
\newblock In \emph{Proceedings of the 58th Annual Meeting of the Association
  for Computational Linguistics}, pages 2453--2470.

\bibitem[{Vaswani et~al.(2017)Vaswani, Shazeer, Parmar, Uszkoreit, Jones,
  Gomez, Kaiser, and Polosukhin}]{vaswani2017attention}
Ashish Vaswani, Noam Shazeer, Niki Parmar, Jakob Uszkoreit, Llion Jones,
  Aidan~N Gomez, Lukasz Kaiser, and Illia Polosukhin. 2017.
\newblock Attention is all you need.
\newblock In \emph{NIPS}.

\bibitem[{Wang et~al.(2018)Wang, Tang, and He}]{wang2018attention}
Yufan Wang, Li~Tang, and Tingting He. 2018.
\newblock Attention-based cnn-blstm networks for joint intent detection and
  slot filling.
\newblock In \emph{Chinese Computational Linguistics and Natural Language
  Processing Based on Naturally Annotated Big Data}, pages 250--261. Springer.

\bibitem[{Weld et~al.(2021)Weld, Huang, Long, Poon, and Han}]{weld2021survey}
Henry Weld, Xiaoqi Huang, Siqu Long, Josiah Poon, and Soyeon~Caren Han. 2021.
\newblock A survey of joint intent detection and slot filling models in natural
  language understanding.
\newblock \emph{ACM Computing Surveys (CSUR)}.

\bibitem[{Xu and Sarikaya(2013)}]{xu2013convolutional}
Puyang Xu and Ruhi Sarikaya. 2013.
\newblock Convolutional neural network based triangular crf for joint intent
  detection and slot filling.
\newblock In \emph{2013 ieee workshop on automatic speech recognition and
  understanding}, pages 78--83. IEEE.

\bibitem[{Xu et~al.(2020)Xu, Haider, and Mansour}]{xu2020end}
Weijia Xu, Batool Haider, and Saab Mansour. 2020.
\newblock End-to-end slot alignment and recognition for cross-lingual nlu.
\newblock In \emph{Proceedings of the 2020 Conference on Empirical Methods in
  Natural Language Processing (EMNLP)}, pages 5052--5063.

\bibitem[{Yao et~al.(2014)Yao, Peng, Zhang, Yu, Zweig, and Shi}]{yao2014spoken}
Kaisheng Yao, Baolin Peng, Yu~Zhang, Dong Yu, Geoffrey Zweig, and Yangyang Shi.
  2014.
\newblock Spoken language understanding using long short-term memory neural
  networks.
\newblock In \emph{2014 IEEE Spoken Language Technology Workshop (SLT)}, pages
  189--194. IEEE.

\bibitem[{Yao et~al.(2013)Yao, Zweig, Hwang, Shi, and Yu}]{yao2013recurrent}
Kaisheng Yao, Geoffrey Zweig, Mei-Yuh Hwang, Yangyang Shi, and Dong Yu. 2013.
\newblock Recurrent neural networks for language understanding.
\newblock In \emph{Interspeech}, pages 2524--2528.

\bibitem[{Ye et~al.(2021)Ye, Manotumruksa, and
  Yilmaz}]{DBLP:journals/corr/abs-2104-00773}
Fanghua Ye, Jarana Manotumruksa, and Emine Yilmaz. 2021.
\newblock \href {http://arxiv.org/abs/2104.00773} {Multiwoz 2.4: {A}
  multi-domain task-oriented dialogue dataset with essential annotation
  corrections to improve state tracking evaluation}.
\newblock \emph{CoRR}, abs/2104.00773.

\bibitem[{Young et~al.(2022)Young, Xing, Pandelea, Ni, and
  Cambria}]{young2022fusing}
Tom Young, Frank Xing, Vlad Pandelea, Jinjie Ni, and Erik Cambria. 2022.
\newblock Fusing task-oriented and open-domain dialogues in conversational
  agents.

\bibitem[{Yu et~al.(2010)Yu, Wang, and Deng}]{yu2010sequential}
Dong Yu, Shizhen Wang, and Li~Deng. 2010.
\newblock Sequential labeling using deep-structured conditional random fields.
\newblock \emph{IEEE Journal of Selected Topics in Signal Processing},
  4(6):965--973.

\bibitem[{Zhang et~al.(2019)Zhang, Kishore, Wu, Weinberger, and
  Artzi}]{zhang2019bertscore}
Tianyi Zhang, Varsha Kishore, Felix Wu, Kilian~Q Weinberger, and Yoav Artzi.
  2019.
\newblock Bertscore: Evaluating text generation with bert.
\newblock In \emph{International Conference on Learning Representations}.

\bibitem[{Zhao et~al.(2021)Zhao, He, Wang, Li, Mi, Liu, Jiang, Liu, and
  Chen}]{zhao2021unids}
Xinyan Zhao, Bin He, Yasheng Wang, Yitong Li, Fei Mi, Yajiao Liu, Xin Jiang,
  Qun Liu, and Huanhuan Chen. 2021.
\newblock Unids: A unified dialogue system for chit-chat and task-oriented
  dialogues.
\newblock \emph{arXiv e-prints}, pages arXiv--2110.

\end{thebibliography}

%\fi

%\appendix
\section{Appendix}
\label{sec:appendix}

Table \ref{tab: use cases study to show the ability to generate transition sentences to enable the system-initiated transitions from chit-chat to task-oriented services.} shows several dialogue examples with generated transition sentences in this work.
Table \ref{tab: transition sentences templates for different domains} and Table \ref{tab: transition sentences templates for different domain-slot pairs} show transition sentence templates for different domains and domain-slot pairs, respectively.

\begin{table*}
\centering
\footnotesize
\begin{tabular}{ll}
    \toprule
    
    \makecell[c]{dialogue \\ context}  & \makecell[l]{\textbf{User:} A group of my friends from another university is coming to visit my campus! \\
    \textbf{System:} What is the occasion? \\
    \textbf{User:} My friends are actually working together to film a short movie and they would \\ like shoot some footage here. \\
    \textbf{System:} It seems your friends are working on quite an interesting project. \\
    \textbf{User:} I am going to meetup with them this Saturday before bringing them to my university.}\\
    
    \midrule
    
    \makecell[c]{response at \\ transition turn} & \makecell[l]{
    \emph{[d]:} That is a good opportunity to catch up with your friends. \\ \textit{\textcolor{red}{By the way, I can help you with the \textbf{train}.}}  \\ 
    \emph{[d-v]:} That sounds like a great opportunity to discuss the film with your friends. \\ \textit{\textcolor{red}{By the way, i am happy to look for a \textbf{train} on \textbf{Saturday} for you.}}} \\
    
    \midrule
    \midrule
    
    \makecell[c]{dialogue \\ context}  & \makecell[l]{\textbf{User:} My sister likes to watch musical plays. \\
    \textbf{System:} Nice! \\
    \textbf{User:} I am thinking of bringing her to a theatre to watch one soon since it will be \\ her birthday next week.}\\
    
    \midrule
    
    \makecell[c]{response at \\ transition turn} & \makecell[l]{
    \emph{[d]:} Sounds fun! \textit{\textcolor{red}{If you are looking for an \textbf{attraction}, I am happy to help.}}  \\ 
    \emph{[d-v]:} Sounds great! \textit{\textcolor{red}{If you are finding any \textbf{theatres}, our \textbf{attraction} service can help.}}} \\

    \midrule
    \midrule
    
    \makecell[c]{dialogue \\ context}  & \makecell[l]{\textbf{User:} I think I have had enough time in Cambridge, its a small town and you really \\ get tired of seeing the same things over and over again . \\
    \textbf{System:} How long have you been here for? \\
    \textbf{User:} I have lived here for about 3 years now, I think I want a change of fresh air.}\\
    
    \midrule
    
    \makecell[c]{response at \\ transition turn} & \makecell[l]{
    \emph{[d]:} That sounds great. \textit{\textcolor{red}{If you want, I can look for a \textbf{train} for you.}}  \\ 
    \emph{[d-v]:} I think change is always good. \textit{\textcolor{red}{if you are planning to travel, I am happy to}} \\ \textit{\textcolor{red}{look for a \textbf{train} leaving from \textbf{Cambridge} for you.}}} \\

    \midrule
    \midrule
    
    \makecell[c]{dialogue \\ context}  & \makecell[l]{\textbf{User:} Without the additional workload, I should being accompany my family on a visit \\ now. Now i have to let my family visit Hobsons house first and meet my family there. \\
    \textbf{System:} Sorry to hear that, hope you can finish the work early. \\
    \textbf{User:} I have finished the work and left the company. I will try to meet my family \\ by bus or taxi, whichever is faster.}\\
    
    \midrule

    \makecell[c]{response at \\ transition turn} & \makecell[l]{
    \emph{[d]:} I am sure your family will have an enjoyable time there. \\ \textit{\textcolor{red}{Shall I get a \textbf{taxi} for you getting there?}}  \\ 
    \emph{[d-v]:} I am sure your family will understand. ,  \textit{\textcolor{red}{By the way, if you want to book}} \\ \textit{\textcolor{red}{a taxi to \textbf{hobsons house}, feel free to use our \textbf{taxi} service.}}} \\

    \bottomrule

\end{tabular}
\caption{Several dialogue examples with transition sentence (highlighted in \textcolor{red}{red}) generated by the extended NLG with TSG. The \emph{[d]} means only the transition domain as transition prompt and \emph{[d-v]} means the transition domain-slot-value as transition prompt to guide the transition sentence generation. Transition domains and values present in transition sentences are highlighted in \textbf{bold}.}
\label{tab: use cases study to show the ability to generate transition sentences to enable the system-initiated transitions from chit-chat to task-oriented services.}
\end{table*}

\begin{table*}
\centering
\footnotesize
\begin{tabular}{ll}
\toprule
domain & templates of the transition sentence \\
\midrule
\makecell[l]{restaurant} & \makecell[l]{
    I am happy to give recommendation on restaurants. \\
    I can recommend some restaurants if you want. \\
    Do you want my recommendation on the restaurants? \\
    I can also provide you more information on this restaurant. \\
    Maybe you would like to use our restaurant service to know more about it. \\
    ...} \\
    
\midrule
\makecell[l]{attraction} & \makecell[l]{
    By the way, you can reach to our attraction service to know more about this place. \\
    Besides, our attraction service provides various information. \\
    I can recommend some attractions to you. \\
    By the way, have you checked out our attraction service to know more about this place? \\
    If you are finding any attraction, I am always happy to help. \\
    ...} \\
    
\midrule
\makecell[l]{train} & \makecell[l]{
    Additionally I could help with looking for train tickets for you. \\
    By the way, I can help you to find thee trains to get there. \\
    Let me arrange the train for you. \\
    Please refer to our train service if you need any help with the booking. \\
    I am glad to give you more information on the train. \\
    ...} \\
    
\midrule
\makecell[l]{taxi} & \makecell[l]{
    Do you need help with booking a taxi to get there? \\
    Do you want me to look for a taxi for you? \\
    Do you need a taxi afterwards? \\
    Maybe you would like my help with the taxi? \\
    If you need to get there soon, I can help you book a taxi. \\
    ...} \\
    
\bottomrule
\end{tabular}
\caption{Transition sentences templates for different domains.}
\label{tab: transition sentences templates for different domains}
\end{table*}

\begin{table*}
\centering
\footnotesize
\begin{tabular}{ll}
\toprule
domain-slot & templates of the transition sentence \\
\midrule
\makecell[l]{restaurant-food} & \makecell[l]{
    I am happy to give recommendation on [VALUE] restaurants. \\
    I can recommend some [VALUE] restaurants if you want. \\
    You can find more information on [VALUE] restaurants in our restaurant service. \\
    It's my pleasure to recommend some [VALUE] restaurants if you want. \\
    ... } \\
    
\midrule
\makecell[l]{restaurant-name} & \makecell[l]{
    I can also provide you more information on this restaurant named [VALUE]. \\
    Maybe you would like to use our restaurant service to know more about [VALUE]. \\
    I will be more than pleasant to help with booking a table at the restaurant called [VALUE]. \\
    Feel free to ask for more information about this restaurant named [VALUE]. \\
    ... } \\
    
\midrule
\midrule
\makecell[l]{attraction-name} & \makecell[l]{
    By the way, you can reach to our attraction service to know more about [VALUE]. \\
    Do you want to plan your trip to [VALUE] using our attraction service? \\
    By the way, I can provide more attraction information on [VALUE]. \\
    Talking about attractions, do you need more information about [VALUE]. \\
    ... } \\

\midrule
\makecell[l]{attraction-type} & \makecell[l]{
    Besides, our attraction service provides various information on [VALUE]. \\
    If you are looking for attraction that has [VALUE] activities, i am happy to help you. \\
    In our attraction service, you can find more information on visiting [VALUE]s. \\
    ...} \\
    
\midrule
\midrule
\makecell[l]{train-day} & \makecell[l]{
    Additionally I could help with looking for train on [VALUE] for you. \\
    Let me arrange the train for [VALUE] for you. \\
    If you want, you can use our service to book the train for [VALUE]. \\
    I would love to help you with the train tickets for [VALUE]. \\
    ... } \\

\midrule
\makecell[l]{train-destination} & \makecell[l]{
    By the way, I can help you to find the trains to [VALUE]. \\
    If you want, I can look for a train to [VALUE] for you. \\
    Additionally, you can use our service to book a train to [VALUE]. \\
    ... } \\

\midrule
\makecell[l]{train-departure} & \makecell[l]{
    I think our service might be helpful in booking the train leaving from [VALUE]. \\
    I am happy to look for a train leaving from [VALUE] for you. \\
    Shall I find you some train tickets departing from [VALUE]. \\
    ...} \\
    
\midrule
\midrule
\makecell[l]{taxi-departure} & \makecell[l]{
    By the way, do you need help with booking a taxi departing from [VALUE]? \\
    Do you want me to look for a taxi depart from [VALUE] for you?. \\
    Will you need my help with the taxi leaving from [VALUE]. \\
    ...} \\

\midrule
\makecell[l]{taxi-destination} & \makecell[l]{
    Shall I get a taxi for you to get to [VALUE]? \\
    By the way, if you need a taxi to [VALUE], please feel free to use our taxi service. \\
    If you need a taxi to get to [VALUE], feel free to use our taxi service. \\
    ...} \\
    
\bottomrule
\end{tabular}
\caption{Transition sentences templates for different domain-slot pairs. The specific values in the human augmented transition sentences are replaced by the special [VALUE] token to collect the templates.}
\label{tab: transition sentences templates for different domain-slot pairs}
\end{table*}

\end{document}